\documentclass{IEEEtran}
\usepackage{cite}
\usepackage{amsmath,amssymb,amsfonts}
\usepackage{multirow}
\usepackage{booktabs}
\usepackage{algorithm}
\usepackage{algorithmic}
\usepackage{graphicx}
\PassOptionsToPackage{square,numbers,sort&compress}{natbib}
\usepackage{threeparttable}
\usepackage{textcomp}
\usepackage{color}
\usepackage[table, dvipsnames]{xcolor}
\usepackage[colorlinks=true, citecolor=gray, linkcolor=teal]{hyperref}       
\def\BibTeX{{\rm B\kern-.05em{\sc i\kern-.025em b}\kern-.08em
    T\kern-.1667em\lower.7ex\hbox{E}\kern-.125emX}}
\begin{document}
\title{Modeling Task Relationships in Multi-variate Soft Sensor with Balanced Mixture-of-Experts}
\author{Yuxin Huang, Hao Wang, Zhaoran Liu, Licheng Pan, Haozhe Li, Xinggao Liu \IEEEmembership{Member, IEEE}
\thanks{The authors are with the State Key Laboratory of Industrial Control Technology, College of Control Science and Engineering, Zhejiang
University, Hangzhou 310027, China (e-mail: horthy123@163.com;
22032130@zju.edu.cn; 22032057@zju.edu.cn; 22132045@zju.edu.cn;
lihaozhe@zju.edu.cn; lxg@zju.edu.cn).}}

\maketitle

\begin{abstract}
Accurate estimation of multiple quality variables is critical for building industrial soft sensor models, which have long been confronted with data efficiency and negative transfer issues. Methods sharing backbone parameters among tasks address the data efficiency issue; however, they still fail to mitigate the negative transfer problem. To address this issue, a balanced Mixture-of-Experts (BMoE) is proposed in this work, which consists of a multi-gate mixture of experts (MMoE) module and a task gradient balancing (TGB) module. The MoE module aims to portray task relationships, while the TGB module balances the gradients among tasks dynamically. Both of them cooperate to mitigate the negative transfer problem. Experiments on the typical sulfur recovery unit demonstrate that BMoE models task relationship and balances the training process effectively, and achieves better performance than baseline models significantly.

\end{abstract}


\section{Introduction}
\label{sec:introduction}
Process industry plays an important role in modern industry and is closely related to key industrial manufacturing such as oil, gas, rare metals, iron, and steel, which forms an integral part of modern human life and national economies. Monitoring dynamic changes of components precisely in the process industry has become one of the main concerns in order to meet urgent but harsh demands such as increasing production, reducing materials consumption, protecting environments, and ensuring safety in manufacturing process. However, in process engineering, there are many difficult-to-measure variables that are critical to evaluate process quality. For example, in deep water gas-lift oil well process, downhole pressure carries useful information for the oil refinery process of oil field \cite{1}. The pressure is hard to be consistently measured by hardware sensors like permanent downhole gauges because the measurement will break down under high pressure and salinity environment \cite{3}.

Soft sensors, which aim to measure these variables in an indirect manner, can be grouped into model-driven and data-driven methods. With the development of machine learning and database technologies~\cite{fanlearnable,6,chen2022directed}, data-driven approaches dominate the field of soft sensors, which construct statistical estimands of quality variables with measurable process variables \cite{4}. Statistical models were first applied to soft sensors, based on principal component regression (PCR), partial least squares (PLS) \cite{6} and distribution mixture models~\cite{dai2022incremental,dai2023variational,dai2020incremental}. To depict nonlinear relationships between variables, soft sensors based on advanced machine learning were further witnessed, such as support vector regression (SVR) \cite{7}, Extreme Learning Machine (ELM) \cite{9}. 
For example, Zhang et al. \cite{10} proposed a Double-level Locally Weighted Extreme Learning Machine (DLWELM) based soft sensor for online prediction of quality variables, which shows better performance compared to conventional statistical methods.

More recently, researchers have turned their attention to deep learning based soft sensors \cite{11,chen2021knowledge}, which map raw data into low-dimensional semantic space to extract spatial/temporal characteristics. Specifically, Recurrent Neural Networks (RNN) and Temporal Convolutional Networks (TCNs)~\cite{13} have been deeply applied to depict temporal dependencies; Convolutional Neural Networks (CNNs) have been widely used to depict spatial dependencies~\cite{14,15}. Autoencoders (AEs) \cite{16} have been widely used to depict dependencies among process variables. For example, Yuan et al. \cite{17} proposed variable-wise weighted stacked autoencoder (VW-SAE) to acquire output-related representations. Sun et al. \cite{18} proposed Gated Stacked Autoencoder (GSAE), which utilized multi-scale representations from multiple layers. 

Although accurate soft sensors for the individual quality variable have been achieved, there are always multiple quality variables that need to be measured simultaneously in the industrial process, namely multi-variate soft sensor (MVSS). For example, to monitor the separation energy consumption and product purity in reactive distillation process, the product concentration and reactant concentration require to be measured simultaneously. Approaches mentioned above, which model each quality variable independently, ignore the correlation between quality variables and thus utilize data in an inefficient way. 

Multi-task learning has long been used to enhance the data efficiency \cite{li2022novel,5,wangescm} and generalization~\cite{liu2023novel} of neural networks.
According to the parameter sharing strategy, it can be categorized into two groups.
Methods in the first group share parameters in a hard approach, such as UberNet \cite{19} and Multilinear relationship networks \cite{20}. This strategy does enhance data efficiency; however, the task relationship is not depicted, suffering from the risk of negative transfer between conflicting tasks.
Methods in the second group share parameters in a soft approach, such as Cross-stitch Networks \cite{22}, Sluice networks \cite{23}, and MTAN \cite{24}, which alleviate negative transfer by depicting tasks relationships. 
Another concern of multi-task soft-sensor is the seesaw phenomenon proposed in \cite{32}. Specifically, in the joint optimization procedure, the optimizer always focuses on the dominant task at the expense of other tasks, making the training process unbalanced.

As such, both lacks of task relationship and the seesaw problem introduce negative transfer, thus hindering the performance of multi-task learning (MTL) models. To alleviate the two problems simultaneously, a balanced Mixture-of-Experts (BMoE) is developed in this work, which consists of a multi-gate mixture of experts (MMoE) module and a task gradient balancing (TGB) module. 
Specifically, the MMoE module finds transferable representations by depicting task relationships; the TGB module addresses the seesaw problem by tuning the gradient magnitudes dynamically. Both of them cooperate to mitigate the negative transfer in MVSS.

The contributions of this paper are summarized as follows:

\begin{enumerate}
    \item {
    A multi variate soft sensor (MVSS) problem is proposed, which aims to estimate multiple quality variables simultaneously. 
    The primary challenge is the negative transfer between quality variables, which is mainly caused by the non-transferable features and seesaw problem.
    }
    \item {
    A BMoE model is proposed to address the negative transfer in MVSS, based on a generalized MMoE module and TGB module.
    Specifically, the MMoE module finds transferable representations by depicting task relationships; and the TGB module addresses the seesaw problem by normalizing the gradient magnitudes among tasks. 
    Both corporate to address the negative transfer.
    }
    \item{
    Extensive experiments are conducted in the sulfur recovery unit process, where BMoE finds the transferable representations efficiently, addresses the seesaw problem effectively, and improves the sensor quality significantly.}
\end{enumerate}

The rest of this paper is organized as follows: Section 2 presents preliminaries on MoE structure and the GradNorm algorithm. Section 3 proposes the novel BMoE approach. Section 4 shows experiments on famous case studies, the sulfur recovery unit. Conclusions are drawn in the final section.

\section{Preliminaries}
\subsection{Multi-task learning models}
\begin{figure}[!t]
\centerline{\includegraphics[width= \columnwidth]{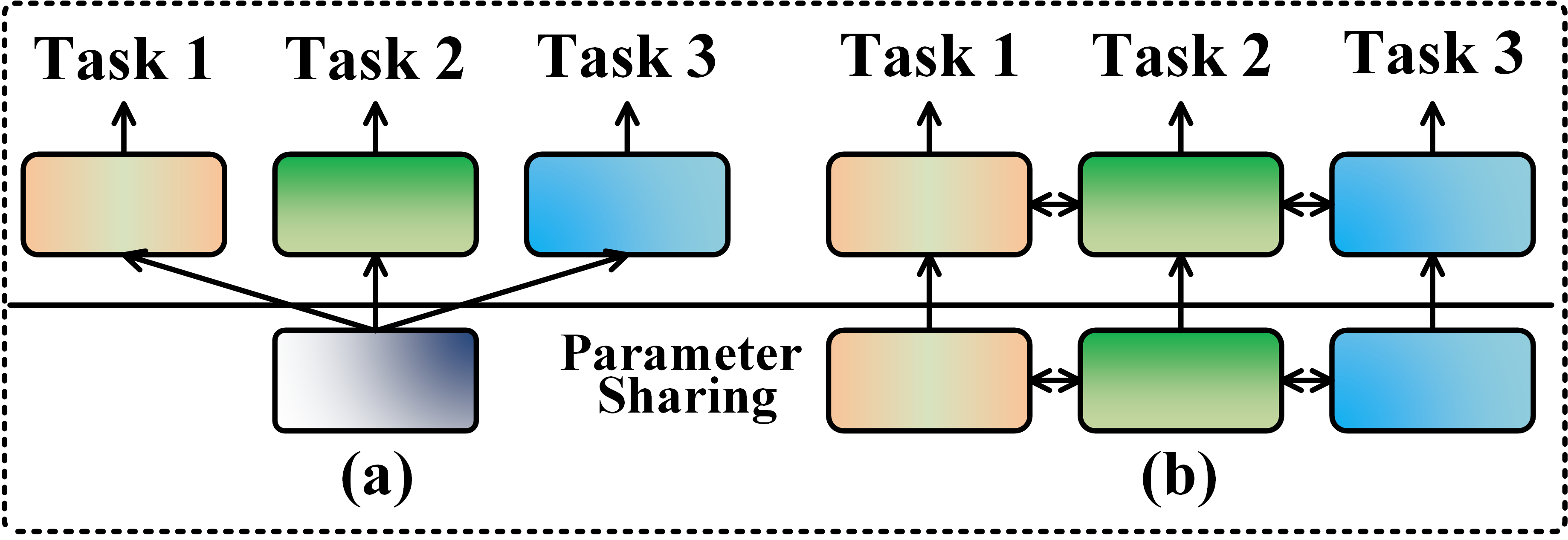}}
\caption{
The structures of MTL models: (a) Hard-share. (b) Soft-share.}
\label{fig2}
\vspace{-0.6cm}
\end{figure}
\indent Multi-task learning aims to improve model generalization by leveraging training signals of multiple tasks. The performance would be improved if the associated tasks share complementary information. Deep multi-task architectures can be categorized into two groups with respect to parameter sharing strategy. In the hard-sharing group, models typically consist of one hard-shared encoder followed by several task-specific heads (see Fig. \ref{fig2} a), which means the hard-shared encoder is forced to be utilized by all different task-specific heads; in the soft-sharing group, the hard-shared encoder is replaced by a feature sharing mechanism that handles the cross-task talk (see Fig. \ref{fig2} b). In that way, parameters of the bottom networks are encouraged to be close by feature sharing mechanisms like loss function, but not forced to be same.
\subsection{Positive \& Negative transfer}
\indent Given $N$ tasks, the loss function of the MTL model can be given as \eqref{MTLLoss} shown, where $\mathcal{B}$ and $\mathcal{A}$ are the share module and specific module, respectively. The learning process seeks to find an MTL model as minimizing \eqref{MTLLoss} over $\mathcal{B}$ and the $\mathcal{A}_i$s.
\begin{equation}
f(\mathcal{A}_1, \mathcal{A}_2, ... ,\mathcal{A}_N;\mathcal{B}) = \sum_{i=1}^{N}{L(g(x_i,\mathcal{B})\mathcal{A}_{i},y_i)}. 
\label{MTLLoss}
\end{equation}
\indent The phenomenon named positive transfer is defined as the model training that through \eqref{MTLLoss} improves over just training the specific task. On the contrary, the negative transfer suggests the training through \eqref{MTLLoss} cannot improve over just training the specific task. In practice, the quality variables of the chemical process may occur the negative transfer phenomenon in the model training. Take the reaction process as an example. To better monitor the reaction, the main reaction product content and the side reaction product are always bound to be monitored simultaneously. However, the product contents measurement tasks will be trade-off results from the selectivity and the capacity. This task indicates that the design of the model should consider the negative transfer phenomenon.
\subsection{Mixture of Expert}
\begin{figure}[htbp]
\centerline{\includegraphics[width= \columnwidth]{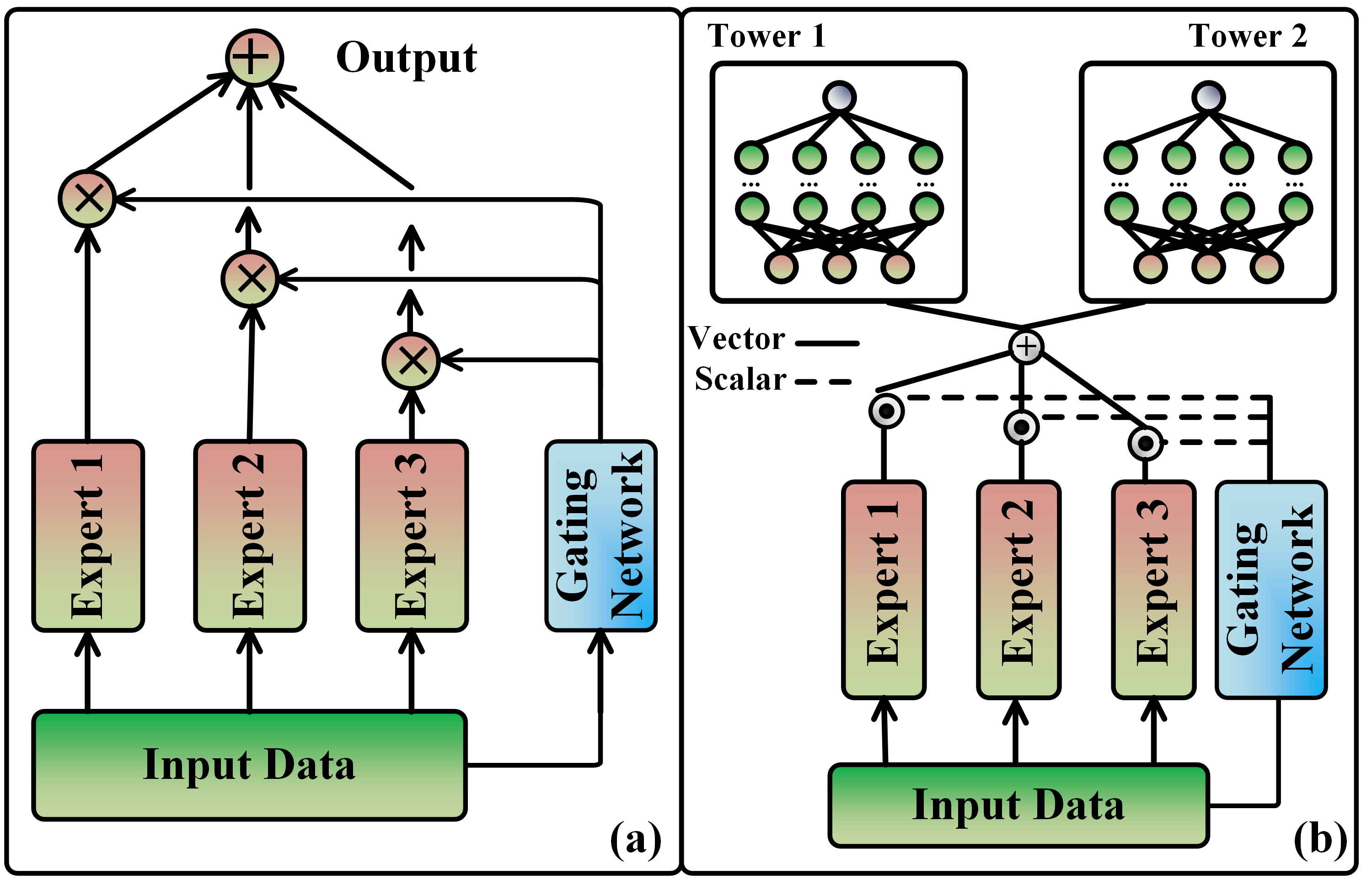}}
\caption{
Structures of (a)  mixture of experts (b) multi-gate mixture of experts.}
\label{MoEandMMoE}
\vspace{-0.3cm}
\end{figure}
\indent Fig. \ref{MoEandMMoE} (a) presents the structure of the mixture of experts (MoE), and \eqref{MoEModel} gives the expression for the MoE.
\begin{equation}
o = \sum_{i=1}^{K}{g_{i}(x) \times f_{i}(x) }, 
\label{MoEModel}
\end{equation}
where the $o$ stands for the output of the model, the $f_{i}(x)$ indicates the output of the corresponding expert, and the $g_{i}(x)$ stands for the weight obtained from the gating network.
Note that, the gating network is designed for calculating the distribution of the corresponding expert, and as such the $g_{i}(x)$ is ought to satisfy the corresponding constraint:
\begin{equation}
\sum_{i=1}^{K}{g_{i}(x)} = 1. 
\end{equation}
\indent Moreover, as shown in Fig \ref{MoEandMMoE} (b), the MoE layer can be introduced into the MTL framework from the perspective of selecting the subset of the expert networks. In one-gated MoE model, the single gating network gives weights of the weighted sum to all expert networks' outputs. On this basis, the weighted sum of features from different experts are then fed to the corresponding networks (known as towers) for specific tasks, which output the prediction results. By adding the MoE layer in the MTL framework, the expert feature can be dynamically weighted for different tasks varying from samples. As such, the negative transfer phenomenon can be alleviated in a way.

\section{Proposed Method}
\subsection{Problem Statement}
\indent Given a size-M set of observed history process variables $[x_1, x_2, ..., x_M] \in \mathbb{R}^{M \times F}$, and the corresponding labels corresponding to size-N task $[y_1^{n}, y_2^{n}, ..., y_M^{n}] \in \mathbb{R}^{M \times N}$.
The task is to predict the labels $[y_{M+1}^{n}, y_{M+2}^{n}, ..., y_{M+H}^{n}] \in \mathbb{R}^{H \times N}$
with known process variables $[x_M+1, x_M+2, ..., x_{M+H}] \in \mathbb{R}^{M \times H}$. \\
\indent Proposed method is designed based on following concerns: 
\begin{enumerate}
\item{The backbone architecture needs to be compact and able to extract the abstract semantics beyond the input data.} 
\item{To mitigate the negative transfer problem, the model ought to depict the task relationships and balance the training procedure.} 
\end{enumerate}
\subsection{Multi-gate Mixture-of-Experts Module}

\begin{figure}[h]
  \centering
  \vspace{-0.3cm}
  \includegraphics[scale=0.40]{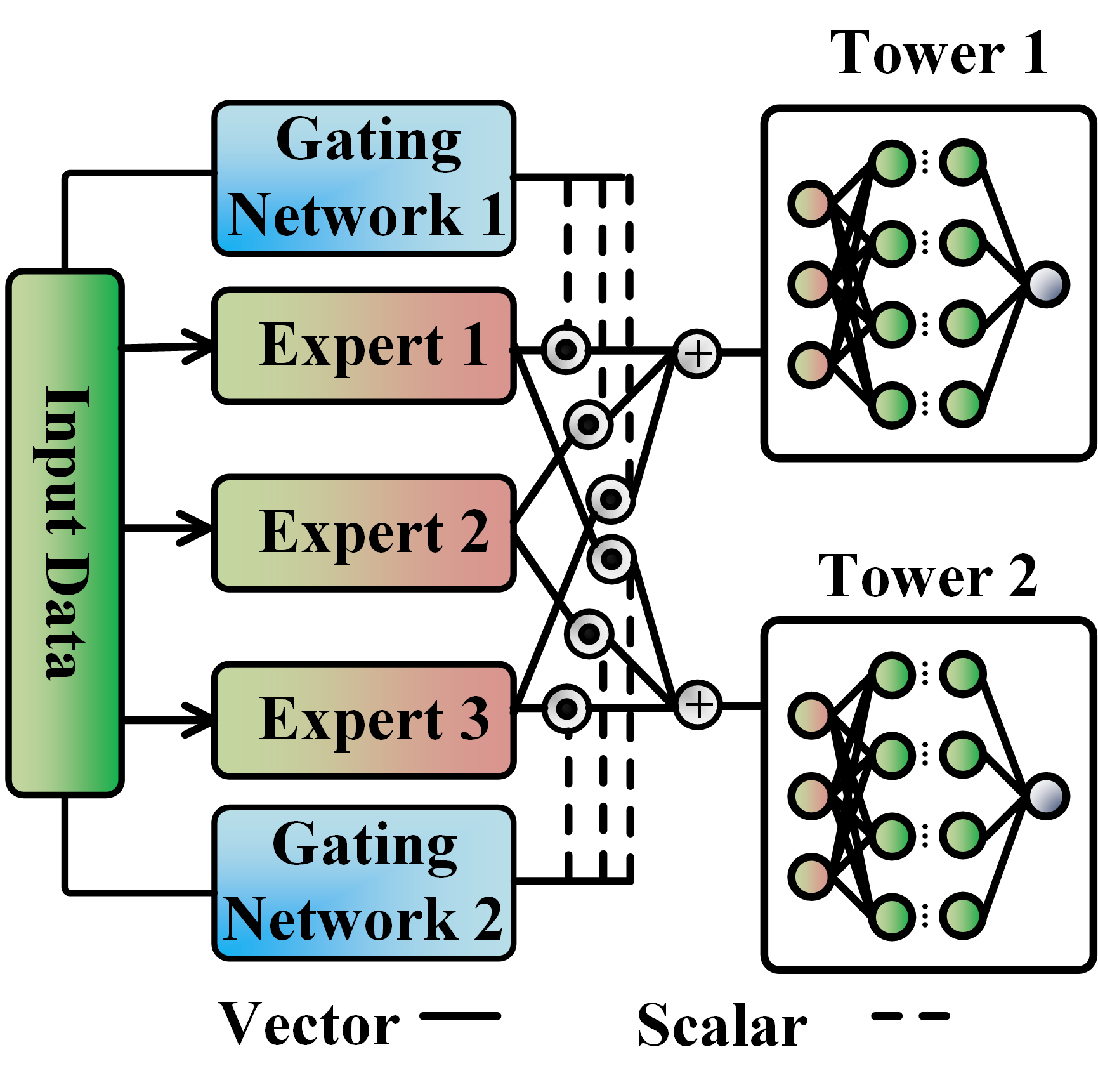}
  \vspace{-0.4cm}
  \caption{The structures of Multi-gate Mixture of Experts.}
\label{figMMoESingleMMoE}
  \vspace{-0.3cm}
\end{figure}
\indent The one-gated MoE network does not model the difference between tasks explicitly, and the negative transfer issue therefore remains unsolved. To this end, inspired by \cite{25}, the Multi-gate Mixture of Experts (MMoE) model is designed in the proposed framework. As shown in Fig. \ref{figMMoESingleMMoE}, raw data is firstly fed into an embedding layer as:
\begin{equation}
\label{eq:embed}
    z=f(x;\theta),
\end{equation}
where $\theta$ is the embedding look-up table hardly shared across tasks. Afterwards, K expert networks $\phi_\mathrm{k}, k=1,2,...,K$ are utilized to extract representations that are softly shared across tasks. Then,
to estimate the i-th quality variable, the experts' outputs are weighted and summed using the gating networks:
\begin{equation}
\label{MMOEOutputExpress2}
\psi_\mathrm{i}= \sum_{\mathrm{k}=1}^Kg_\mathrm{i}^\mathrm{k}(z)_\mathrm{k}\phi_\mathrm{k}(z),\quad i=1,2,...,N,
\end{equation}
where $g_\mathrm{i}^\mathrm{k}(z)$ denotes the gating network of the k-th expert to estimate the i-th quality variable. Finally, the estimate of the i-th quality variable is given as 
\begin{equation}
\label{eq:predict}
    \widehat{y}_\mathrm{i}=\Phi_\mathrm{i}(\psi_\mathrm{i}),\quad i=1,2,...,N,
\end{equation}
where $\Phi_\mathrm{i}$ denotes the task-specific tower function. Leveraging the task-specific tower functions and softly shared experts, task-specific features are extracted and the negative transfer is remitted.

\indent Even though the structure of the gating network in the MMoE is similar to the MoE model,

there still remain differences. Particularly, MoE model equips one gate merely for different tasks, which indicates that the gate needs to serve all the relevant towers. While the MMoE model equips each task with a gating network. On this basis, the gating network would activate the best expert for the specific variable, and generate distributions varying the tasks. To this extent, the negative transfer phenomenon can be mitigated in a way.  

\subsection{Gradient Balance Module}

Another challenge to multi-variate soft sensors owns to the risk of imbalanced training. Specifically, training shared parameters in multi-variate prediction structure is not always balanced for different tasks, with the optimizer overly focusing on the dominant task at the expense of other tasks. To this end, the gradient balance module \cite{26} is designed to balance the gradient magnitudes by adjusting the weights of tasks dynamically. 

Generally, the loss function for multi-variate soft sensor is:
\begin{equation} L = \sum_{\mathrm{i}=1}^{N}w_\mathrm{i}L_\mathrm{i}(t),\end{equation}
where $L_\mathrm{i}(t)$ is the prediction loss of the $i$-th quality variable at the time t, $w_\mathrm{i}$ is the corresponding weight. 
However, the fixed weights are hard to decide, requiring dedicated parameter tuning, especially when the number of quality variables increases. Meanwhile, the unchangeable weight tends to make the training process suffer from training imbalance. Therefore, the dynamic weights that fluctuate over time are introduced.
\begin{equation} L(t) = \sum_{\mathrm{i}=1}^{N}w_\mathrm{i}(t)L_\mathrm{i}(t).\end{equation}

Let $W$ be the parameters that are hardly-shared across tasks, \textit{i}.\textit{e}., the embedding layer in (\ref{eq:embed}). 
The gradient's norm of $w_\mathrm{i}(t)L_\mathrm{i}(t)$ with respect to $W$ is
\begin{equation} 
G^{(\mathrm{i})}_W(t) = ||\bigtriangledown_Ww_\mathrm{i}(t)L_\mathrm{i}(t)||_2,
\end{equation}
which measures the gradient magnitude of the i-th task. 

Forcing gradient magnitudes equal ignores the difference across tasks.
Generally, tasks with higher loss magnitudes ought to dominate the weight updating process.
As such, the relative inverse training rate (RITR) is introduced.
Specifically, defining a baseline magnitude of loss function at time t as
\begin{equation} \widetilde{L}_\mathrm{i}(t) = L_\mathrm{i}(t) / L_\mathrm{i}(0),\end{equation}
and the RITR w.r.t the i-th task is
\begin{equation} r_\mathrm{i}(t) = \frac{\widetilde{L}_\mathrm{i}(t)}{\mathbb{E}[\widetilde{L}_\mathrm{i}(t)]}, \end{equation}
\begin{equation} \mathbb{E}[\widetilde{L}_\mathrm{i}(t)] =\sum_{\mathrm{i}=1}^{N}\widetilde{L}_\mathrm{i}(t) / {N}. \end{equation}

Afterwards, magnitudes of the i-th task $G^{(\mathrm{i})}_W(t)$ is adjusted to $\Bar{G}_W(t) \times {[r_\mathrm{i}(t)]}^\alpha$, where $\alpha$ is a hyperparameter needs to be tuned. Specifically, $\alpha$ sets the force that pulls different tasks back to a common training balance. When the $\alpha$ is higher, it encourages update of networks' weights depends more on the tasks that produce relatively high losses. Then, the discrepancy between $G^{(\mathrm{i})}_W(t)$ and $\Bar{G}_W(t) \times {[r_\mathrm{i}(t)]}^\alpha$ for all tasks is
\begin{equation} L_\mathrm{grad}(t;w_\mathrm{i}(t)) = \sum_\mathrm{i} \left\| G^{(\mathrm{i})}_W(t)-\Bar{G}_W(t) \times {[r_\mathrm{i}(t)]}^\alpha \right\|_1.\end{equation}

The target gradient norm $\Bar{G}_W(t) \times {[r_\mathrm{i}(t)]}^\alpha$ is treated as constant when differentiating this loss function to avoid the loss weights $w_\mathrm{i}(t)$ dropping to zero. And finally, the weight $w_\mathrm{i}(t), i=1,2,…,N$ is updated by:
\begin{equation} w(t+1) = w(t) + \lambda \bigtriangledown_{w_\mathrm{i}}L_\mathrm{grad}. \label{ParamsUpdate}\end{equation}
 
\subsection{Architecture of Balanced Mixture-of-Experts}
\begin{figure}[h]
  \centering
  \vspace{-0.3cm}
  \includegraphics[scale=0.28]{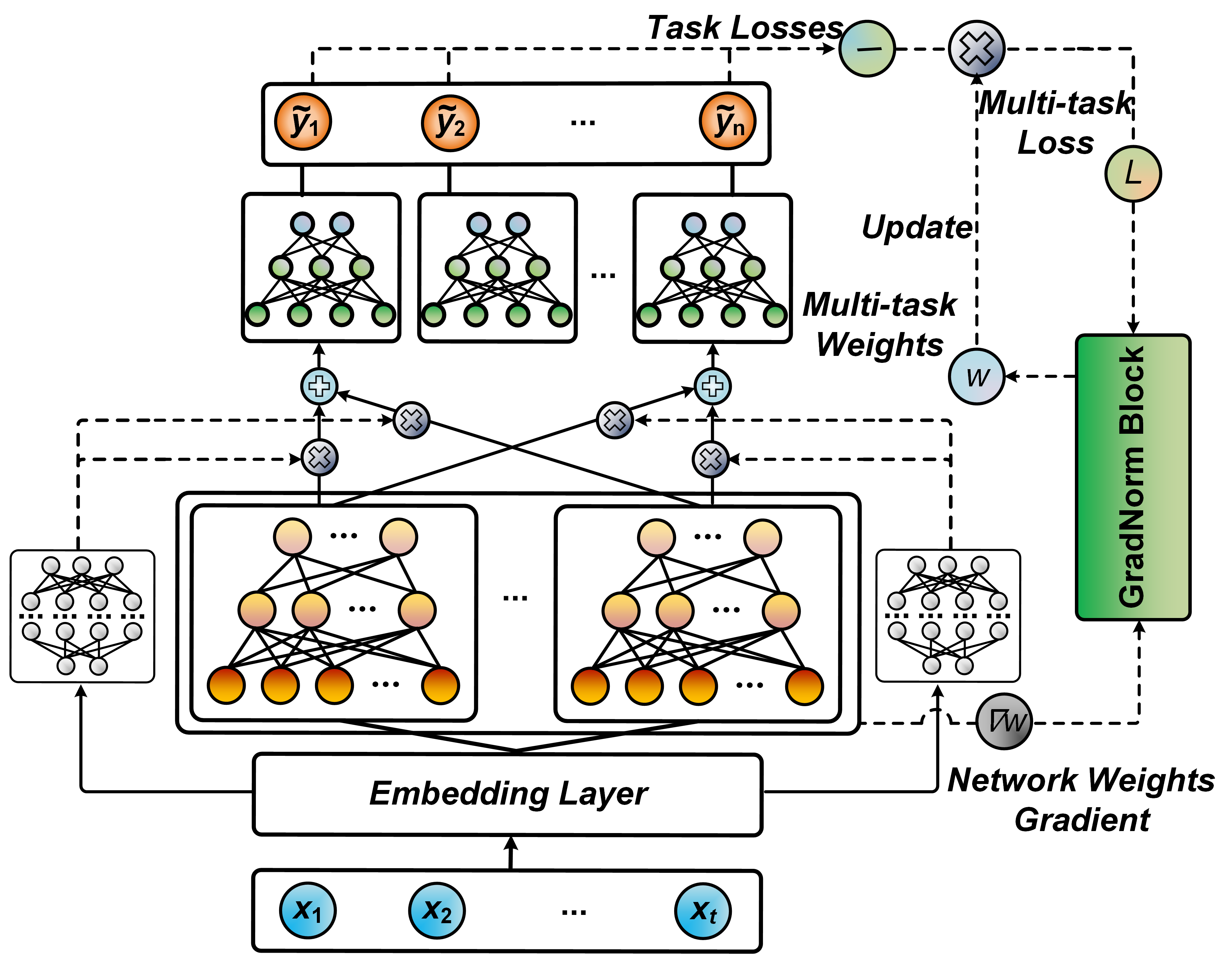}
  \vspace{-0.3cm}
  \caption{The structures of proposed Balanced Mixture of Experts.}
\label{figMMoESingle}
  \vspace{-0.6cm}
\end{figure}

Fig. \ref{figMMoESingle} shows the architecture of Balanced Mixture-of-Experts, which is composed of MoE module and TGB module. For the MoE module, the embedding layer embeds inputs into a semantic representation, which is fed into gating networks and expert networks. The output of them are aggregated as per Eq. (\ref{MMOEOutputExpress2}), followed by task-specific towers which predict the corresponding quality variables. 
For the TGB module, every task is endowed with an initial weight value $\omega_\mathrm{i}(0)=1$.
After that, the hyperparameter $\alpha$ needs to be decided. The embedding layer is chosen to be the layer that actually applied GradNorm, from which we extract the weights $W$. Thereafter, in the training process, the forward propagation is executed, and loss $L_\mathrm{i}(t)$ for batch $x_\mathrm{i}$ is computed. Consequently, the $G^{\mathrm{i}}_W(t)$, $r_\mathrm{i}(t)$, $L_\mathrm{grad}$, $\bigtriangledown_{w_\mathrm{i}}L_\mathrm{grad}$ are obtained. On this basis, the $w_\mathrm{i}(t)$ is updated according to the \eqref{ParamsUpdate}, while $\Theta(t)$ is updated using the standard backward pass. $w_\mathrm{i}(t)$ also needs to be renormalized to make sure that $\sum_{\mathrm{i}=1}^Nw_\mathrm{i}(t+1) = N$ for $N$ tasks.

Comparing to the widely used Relu in \eqref{ReluFunction}, the mish in \eqref{MishFunction} is designed as the activation function \cite{27}. As such, the model can keep good performance as the network layer increases, thanks to the smoothness brought by the mish, which makes the model receive and spread the information better. Besides, to remit the overfitting phenomenon, the dropout operation that randomly disconnects some parameters by a certain percentage during the training process is adopted. In addition, the dynamic learning rate is also adopted for the framework, which means the learning rate will decay to a much smaller one after certain training epochs.
\begin{equation}
    Relu(x) = \max(0,x). 
    \label{ReluFunction}
\end{equation}
\begin{equation}
    Mish(x) = x \cdot tanh(ln(1+e^x)). 
    \label{MishFunction}
\end{equation}

\subsection{Algorithm Procedure}
\begin{algorithm}[tb]
\caption{BMoE Training Algorithm}
\label{alg:algorithm}
\textbf{Forward Propagation}\\
\textbf{Input}: Sample data $X$ after normalization.\\
\textbf{Parameter}: $T$: max iteration number; $K$: number of expert networks; $f$: function of shared layer; $\phi_\mathrm{k}$ function of $\mathrm{k}$th expert network; $g_\mathrm{i}$ function of gating network with respect to $\mathrm{i}$th variable; $t_\mathrm{i}$ function of the tower network with respect to $\mathrm{i}$th variable.\\
\textbf{Output}: Predictions for variables $\left\{\widehat{y}_\mathrm{i}, i = 1,2,\dots,N   \right\}$

\begin{algorithmic}[1]
\FOR{$t\in[0,\textrm{T}]$}
\FOR{Batch input data $x \in X$}
\STATE $\widehat{x} \leftarrow f(x)$ (through shared layer to reduce dimension of input data).
\STATE Compute $\phi_\mathrm{k}(\widehat{x})$ (through experts layer).
\STATE Compute $ g_\mathrm{i}(\widehat{x}) $ (through gating networks)
\STATE $\psi_\mathrm{i}(\widehat{x}) \leftarrow \sum_{\mathrm{k}=1}^Kg_\mathrm{i}(\widehat{x})_\mathrm{k}\phi_\mathrm{k}(\widehat{x})$; $\widehat{y}_\mathrm{i} \leftarrow \Phi_\mathrm{i}(\psi_\mathrm{i}(\widehat{x}))$
\ENDFOR
\ENDFOR
\end{algorithmic}
\textbf{Backward Propagation}\\
\textbf{Input}: $\alpha$: relative inverse training rate coefficient; $W$: network weights of shared layer; $\omega_\mathrm{i}(0)$: initial task weight for task $\mathrm{i}$; $y_\mathrm{i}$: batch real data for task i; $\widehat{y}_\mathrm{i}$: batch prediction data for task i; $\mathcal{W}$: total network weights\\
\textbf{Parameter}:  $T$: max iteration number; $f_\mathrm{i}$: loss function for task i; $L_\mathrm{i}$: prediction loss of task i; 
$G^\mathrm{(i)}_W$: the $L_2$ norm of the gradient of the weighted single-task loss $\omega_\mathrm{i}(t)L_\mathrm{i}(t)$ with respect to the shared weights $W$;
$\bar{G}_W$: average gradient norm within all tasks; 
$\tilde{L}_\mathrm{i}$: loss ratio for task i;
$r_\mathrm{i}(t)$: relative inverse training rate for task i; $\L_\mathrm{grad}$: $L_1$ loss function between actual and target gradient norms;
$N$: number of tasks; \\
\textbf{Output}: Trained multi-task network.
\begin{algorithmic}[1] 
\FOR{$t\in[0,\textrm{T}]$}
\STATE $L_\mathrm{i}(t) \leftarrow f_\mathrm{i}(y_\mathrm{i},\widehat{y}_\mathrm{i})$; $L(t) \leftarrow \sum_\mathrm{i}\omega_\mathrm{i}(t)L_\mathrm{i}(t)$
\STATE $G^\mathrm{(i)}_W(t) \leftarrow \Vert \nabla_W\omega_\mathrm{i}(t)L_\mathrm{i}(t) \Vert_2$; $\bar{G}_W(t) \leftarrow E_\mathrm{task}[G^\mathrm{(i)}_W(t)]$
\STATE $\tilde{L}_\mathrm{i}(t) \leftarrow \frac{L_\mathrm{i}(t)}{L_\mathrm{i}(0)}$; $r_\mathrm{i}(t) \leftarrow \frac{\tilde{L}_\mathrm{i}(t)}{E_\mathrm{task}[\tilde{L}_\mathrm{i}(t)]}$
\STATE $L_\mathrm{grad} \leftarrow \sum_\mathrm{i}\lvert G^\mathrm{(i)}_W(t) - \bar{G}_W(t) \times [r_\mathrm{i}(t)]^{\alpha} \rvert_1$
\STATE Compute $\nabla_{\omega_\mathrm{i}}L_\mathrm{grad}$, with Keeping targets $\bar{G}_W(t) \times [r_\mathrm{i}(t)]^{\alpha}$ constant
\STATE update $\omega_\mathrm{i}(t+1)$ using $\nabla_{\omega_\mathrm{i}}L_\mathrm{grad}$; update $\mathcal{W}(t+1)$ using $\nabla_{\mathcal{W}}L(t)$
\STATE Normalize $\omega_\mathrm{i}(t+1)$ to make that $\sum_\mathrm{i}\omega_\mathrm{i}(t+1) = N$
\ENDFOR
\end{algorithmic}
\end{algorithm}

Total algorithm procedure is formulated as Algorithm \ref{alg:algorithm}, and the performance of the model is evaluated using root mean square error (RMSE), mean absolute error (MAE), and $\mathrm{R^2}$ listed as below:

\begin{equation}
    RMSE = \sqrt{\frac{1}{M}\sum_{\mathrm{i}=1}^{M}{(y_\mathrm{i} - \widehat{y}_\mathrm{i})}^2}. 
\end{equation}
\begin{equation}
    MAE = \frac{1}{M}\sum^M_{\mathrm{i}=1}|(y_\mathrm{i}-\widehat{y}_\mathrm{i})|. 
\end{equation}
\begin{equation}
    R^2 = 1 - \frac{\sum_{\mathrm{i}=1}^{M}{(y_\mathrm{i}-\widehat{y}_\mathrm{i})}^2}{\sum_{M}{(y_\mathrm{i}-\Bar{y})}^2}. 
\end{equation}
where $M$ is the number of samples; $y_\mathrm{i}, \widehat{y},\Bar{y}$ denote the real value, predicted value, mean value of y, respectively. Generally, lower value of RMSE and MAE indicates lower prediction error of the model, while higher value of $R^2$ score shows a better ability for the model to make predictions. 
\section{Experiments}
In this section, experiments are conducted to evaluate performance of proposed method and answer such research topics: 
\begin{enumerate}
\item{ Does BMoE outperform other approaches significantly;} 
\item{ How does Gradient Balance Module balance training procedure;} 
\item{ Is BMoE robust under different settings of hyperparameters.}
\end{enumerate}

\subsection{Experimental Setup}
\subsubsection{Dataset}
The Sulfur Recovery Unit (SRU) is a vital part in refinery process, which removes pollutants from acid gas streams produced in chemical process to recover sulfur from process \cite{28}. The following reactions will take place in SRU:
\begin{equation}
\begin{cases}
2\mathrm{H_2S} + 3 \mathrm{O}_2 = 2\mathrm{SO_2} + 2\mathrm{H_2O} \\
\frac{2}{3}x\mathrm{H_2S} + \frac{1}{3}x \mathrm{SO_2} = \mathrm{S}_x + \frac{2}{3}x\mathrm{H_2O}
\end{cases}
\end{equation}
The tail gas of SRU still contains $\mathrm{H_2S}$ and $\mathrm{SO_2}$ residuals, which should be monitored before being released into the atmosphere in case such contaminants cause danger to the public. However, the SRU system has features including but not limited to non-linearity and multi-variate. It is difficult to directly measure $\mathrm{H_2S}$ and $\mathrm{SO_2}$ concentration due to the harsh working environment, which is necessary to construct soft sensors of such components. From the perspective of chemical reactions, decreasing from $\mathrm{H_2S}$ concentration may result in increasing of $\mathrm{SO_2}$ concentration from the first reaction. However, from the second reaction, the decrease of $\mathrm{H_2S}$ concentration will decrease the $\mathrm{SO_2}$ concentration. How to establish task relationships between these two components remains a challenge. 

\begin{table}[]{}
\centering
\caption{Relevant variables}
\resizebox{\columnwidth}{!}{
\begin{tabular}{ll}
\hline
Variables & Notations                                                                   \\ \hline
$x_1$      & Gas flow MEA\_GAS                                          \\
$x_2$      & Air flow AIR\_MEA                                                 \\
$x_3$      & Secondary air flow AIR\_MEA\_2                                              \\
$x_4$         & Gas flow in Sour Water Stripping (SWS) zone               \\
$x_5$          & Air flow in SWS zone AIR\_SWS\_TOT                                          \\
$y_1$          & Concentration of $\mathrm{H_2S}$ in the tail gas                                      \\
$y_2$          & Concentration of $\mathrm{SO_2}$ in the tail gas                                \\ \hline
\end{tabular}}
\label{tab1}
\end{table}

\begin{figure}[h]
  \centering
  \vspace{-0.3cm}
  \includegraphics[scale=0.25]{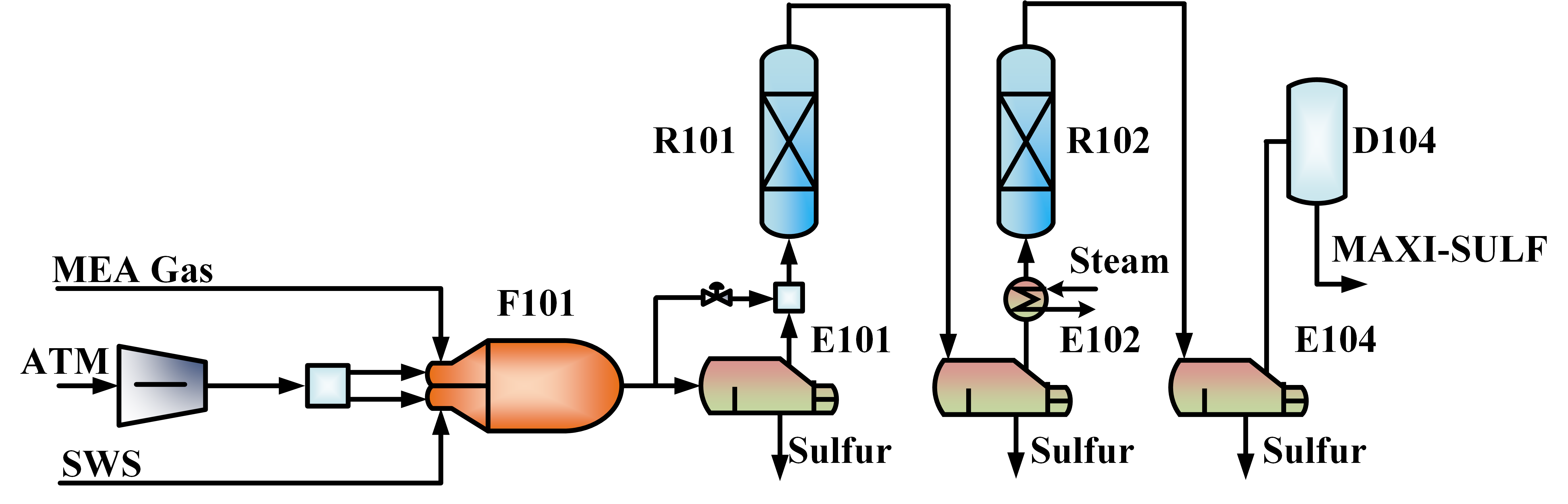}
  \vspace{-0.3cm}
  \caption{The flowsheet of the SRU unit}
\label{SRUUint}
  \vspace{-0.4cm}
\end{figure}

To construct the soft sensor model, five process variables marked in the flowsheet of SRU in Fig. \ref{SRUUint} are selected as the covariates. The detailed descriptions are listed in Table \ref{tab1}.

\subsubsection{Experimental Details}
To better illustrate the superiority of the proposed methods, some single-task learning methods that are widely used in soft sensors are selected to evaluate our proposed model: Partial Least Squares Regression (PLSR), Autoencoder (AE), and its variant, Stacked Autoencoder (SAE), Gated Stacked Autoencoder (GASE) \cite{18}, Long Short-Term Memory Network (LSTM) \cite{29}, Variable Attention-based Long Short-Term Memory (VALSTM) \cite{30}, Supervised Long Short-Term Memory (SLSTM) \cite{31}.

Since these models were originally designed for uni-variate prediction, which is incompatible with the motivation of this work. For the concern of fairness, the multi-variate variants are also proposed in the result comparison. In the single-task learning models, each quality variable is modeled via one model, while in the multi-task learning models multiple quality variables are modeled by one model.

The data is divided with a ratio of 6:2:2 for training, validation, and test. Afterward, normalization is applied for all the subsets separately. All models are trained with a learning rate of 0.01, and a weight decay rate of 0.001. All experiments are proposed on a desktop with Intel i9 Gen, Nvidia RTX 2060 under Python 3.8. The PLS model is constructed on the Sklearn package of Python, and the deep learning models are constructed on PyTorch. All experiments are run at least ten times under ten random seeds. 

\subsection{Performance comparison}

\begin{table*}[]
\caption{Predictive performance (mean$\pm$std) for $\mathrm{H_2S}$ and $\mathrm{SO_2}$.}
\renewcommand{\arraystretch}{1.5}
\resizebox{\textwidth}{!}{
\begin{threeparttable}
\begin{tabular}{ccccccccc}
\hline
 & & \multicolumn{3}{c}{\(\text{H}_{2}\text{S}\)} & & \multicolumn{3}{c}{\(\text{SO}_{2}\)}   \\
 & Models& RMSE & MAE & \(R^{2}\) & & RMSE & MAE & \(R^{2}\) \\
 \hline
\multirow{7}{13pt}{ST} & PLS & 0.0347\(\pm\)0.0002 & 0.0207\(\pm\)0.0001 &
0.6236\(\pm\)0.0033 & & 0.02\(68 \pm\)0.0000 & 0.0\(206 \pm\)0.00002 &
0.7867\(\pm\)0.0007 \\
& AE & 0.0333\(\pm\)0.0008 & 0.0236\(\pm\)0.0020 & 0.6547\(\pm\)0.0167 &
& 0.0\(266 \pm\)0.0011 & 0.0\(212 \pm\)0.0009 & 0.7895\(\pm\)0.0170 \\
& SAE & 0.0334\(\pm\)0.0007 & 0.0242\(\pm\)0.0009 & 0.6517\(\pm\)0.0137
& & 0.0\(265 \pm\)0.0010 & 0.0\(210 \pm\)0.0008 & 0.7917\(\pm\)0.0163 \\
& GSAE & 0.0286\(\pm\)0.0008 & 0.0190\(\pm\)0.0013 & 0.7442\(\pm\)0.0148
& & 0.025\(5 \pm\)0.0006 & 0.01\(97 \pm\)0.0006 & 0.8063\(\pm\)0.0086 \\
& LSTM & 0.0290\(\pm\)0.0004 & 0.0178\(\pm\)0.0006 & 0.7368\(\pm\)0.0081
& & 0.02\(61 \pm\)0.0004 & 0.01\(97 \pm\)0.0003 & 0.7980\(\pm\)0.0067 \\
& VALSTM & 0.0278\(\pm\)0.0010 & 0.0178\(\pm\)0.0009 &
0.7568\(\pm\)0.0165 & & 0.0268\(\pm\)0.0007 & 0.0207\(\pm\)0.0007 &
0.7847\(\pm\)0.0111 \\
& SLSTM & {\underline{0.0231\(\mathbf{\pm}\)0.0003}} &
{\underline{0.0162\(\mathbf{\pm}\)0.0007}} &
{\underline{0.8321\(\mathbf{\pm}\)0.0040}} & &
{\underline{0.0216\(\mathbf{\pm}\)0.0006}} &
{\underline{0.0152\(\mathbf{\pm}\)0.0007}}
&
{\underline{0.8604\(\mathbf{\pm}\)0.0075}} \\
\hline
\multirow{4}{13pt}{MT} & GSAE & 0.0419\(\pm\)0.0038 & 0.0310\(\pm\)0.0029 &
0.4476\(\pm\)0.1035 & & 0.0274\(\pm\)0.0007 & 0.0213\(\pm\)0.0009 &
0.7774\(\pm\)0.0118 \\
& VALSTM & 0.0290\(\pm\)0.0016 & 0.0180\(\pm\)0.0011 &
0.7357\(\pm\)0.0289 & & 0.0300\(\pm 0.0012\) & 0.0232\(\pm\)0.0008 &
0.7296\(\pm\)0.0206 \\
& SLSTM & 0.0253\(\pm\)0.0004 & 0.0199\(\pm\)0.0005 &
0.7983\(\pm\)0.0071 & & 0.0303\(\pm\)0.0003 & 0.0268\(\pm\)0.0002 &
0.7251\(\pm\)0.0051 \\
& MMoE & 0.0234\(\pm\)0.0006 & 0.0165\(\pm\)0.0009 & 0.8286\(\pm\)0.0093
& & 0.0229\(\pm\)0.0010 & 0.0181\(\pm\)0.0005 & 0.8440\(\pm\)0.0142 \\
\multirow{1}{13pt}{\textbf{Ours}} & \textbf{BMoE} &
\textbf{0.0207}\(\mathbf{\pm}\)\textbf{0.0013*} &
\textbf{0.0149}\(\mathbf{\pm}\)\textbf{0.0007*} &
\textbf{0.8665}\(\mathbf{\pm}\)\textbf{0.0163*} & & \textbf{0.0222\(\pm\)0.0006}
& \textbf{0.0176\(\pm\)0.0006} & \textbf{0.8534\(\pm\)0.0079} \\
\bottomrule
 \vspace{-0.8cm}
\end{tabular}
\end{threeparttable}
}
\label{EvaluationTable}
\end{table*}

\begin{figure*}[!t]
\centerline{\includegraphics[scale = 0.45]{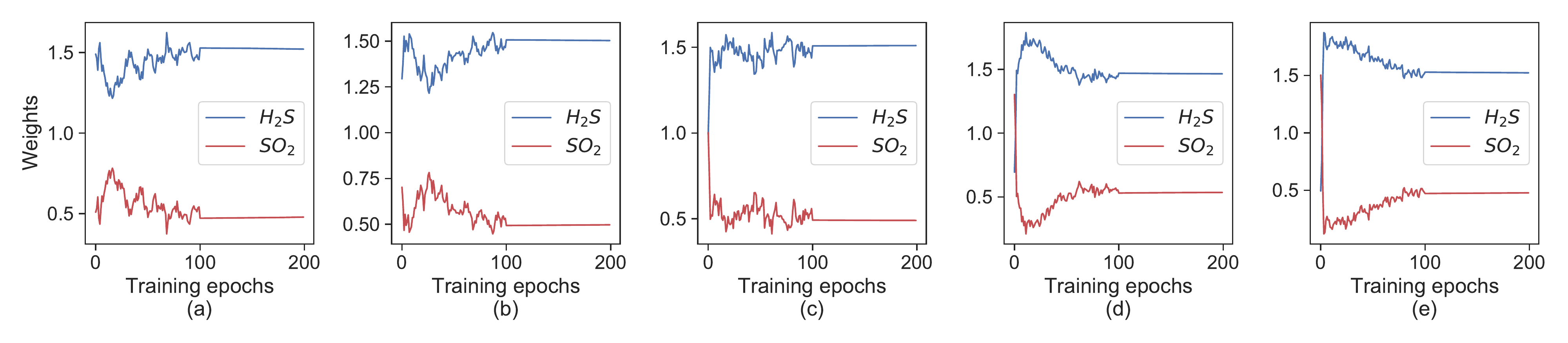}}
\vspace{-0.6cm}
\caption{
The dynamic change of task weights adjusted by the GradNorm block with different intial task weightes. (a) (1.5, 0.5), (b) (1.3, 0.7), (c) (1.0, 1.0), (d) (0.7, 1.3), (e) (0.5, 1.5) for ($\mathrm{H_2S}$, $\mathrm{SO_2}$).}
\vspace{-0.6cm}
\label{fig6}
\end{figure*}

\begin{figure}[!t]
\centerline{\includegraphics[width=\columnwidth]{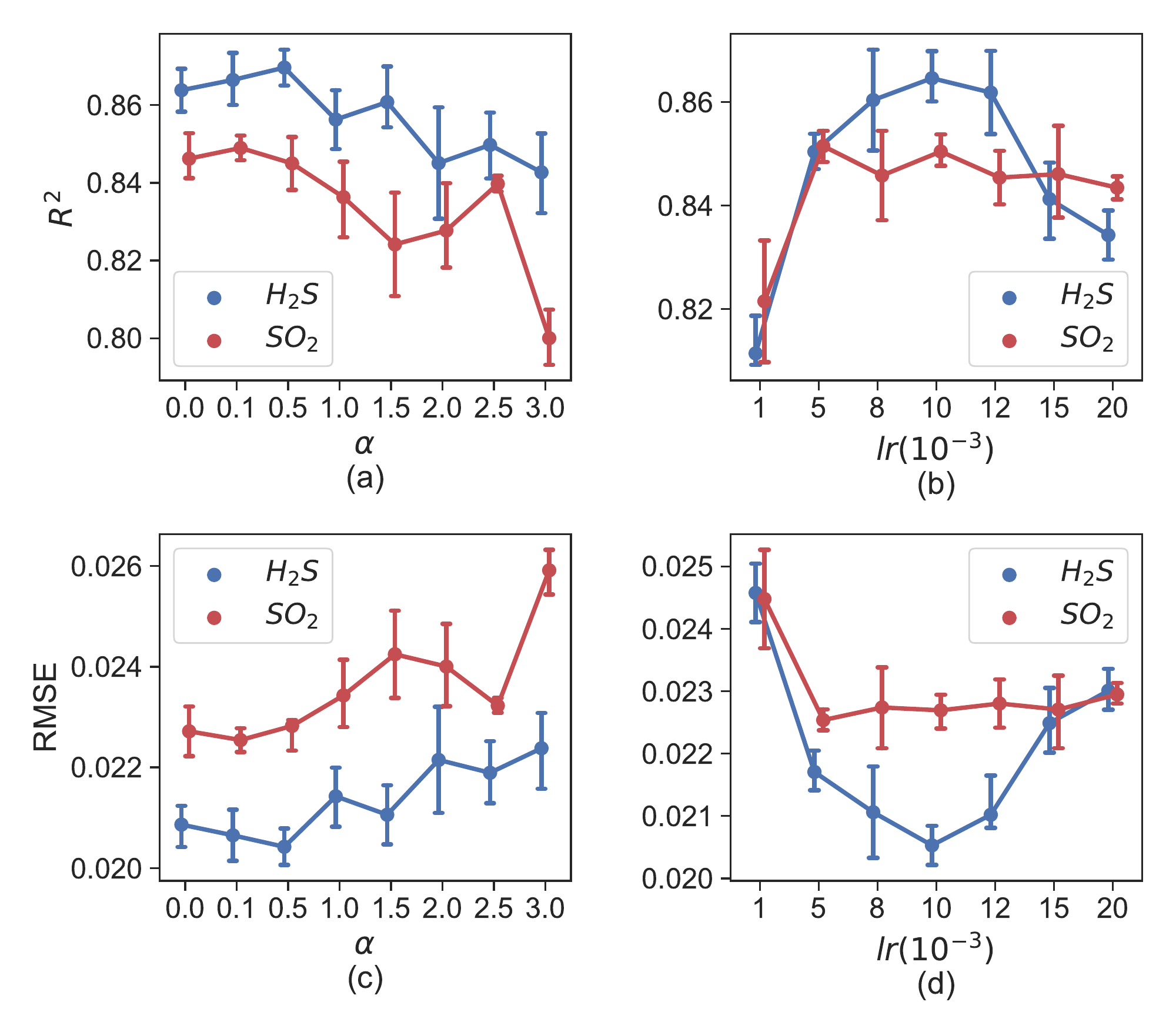}}
\vspace{-0.6cm}
\caption{
Parameter study of the relative inverse training rate coefficient $\alpha$ and the learning rate $lr$.}
\label{fig7}
\vspace{-0.6cm}
\end{figure}

\begin{figure}[!t]
\centerline{
  \includegraphics[width=\columnwidth]{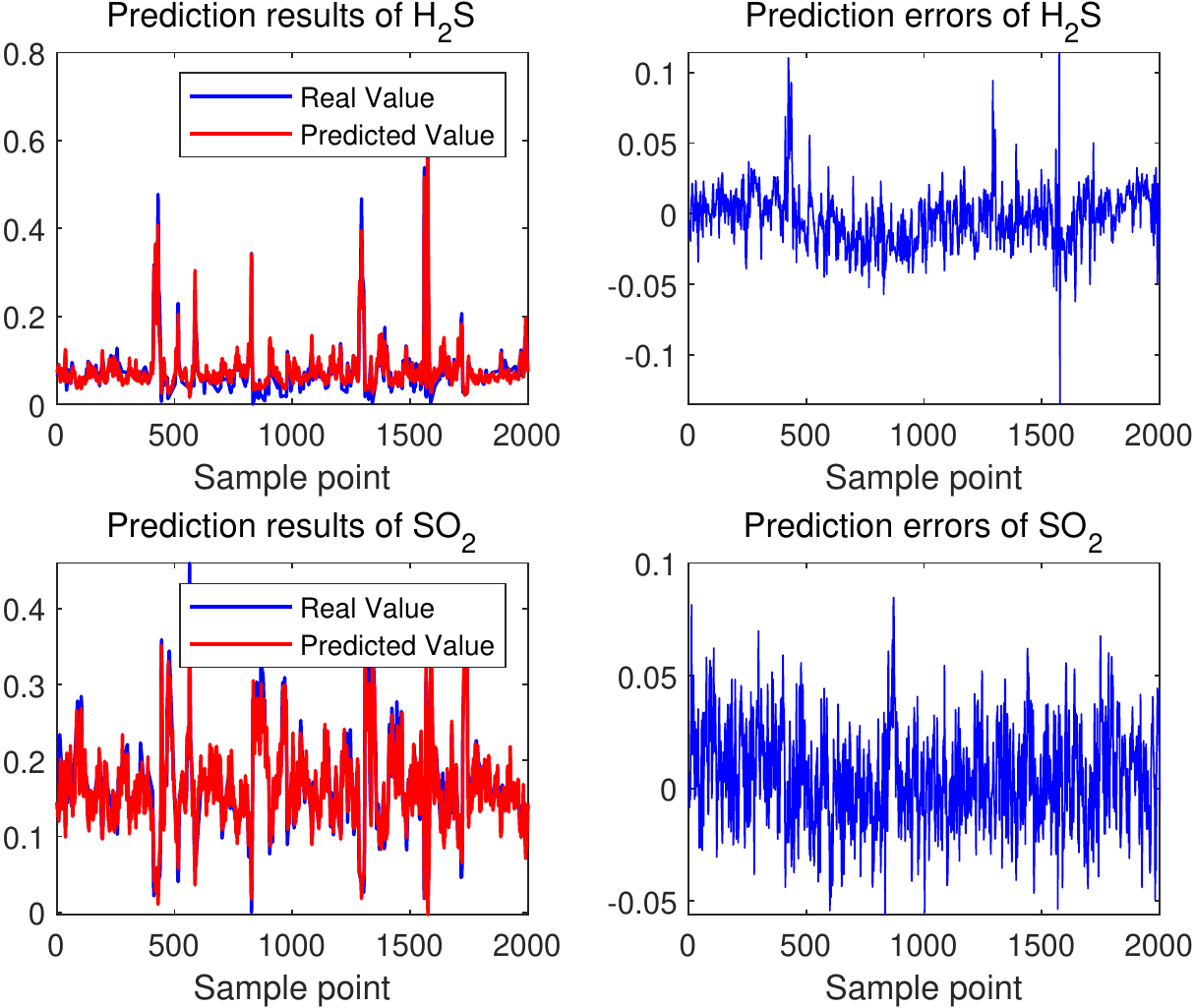}
  }
\vspace{-0.3cm}
\caption{
Prediction results on SRU benchmark.}
\label{fig4}
\vspace{-0.6cm}
\end{figure}

The comparison between the performance of the proposed BMoE and that of its baseline methods is mainly considered in this section, which is conducted on the SRU benchmark, with mean and standard deviation reported over ten runs with different random seeds. The numerical indices are listed in Table II. Underlined results indicate the best baselines over each metric. "*" marks the methods that improve the best baselines significantly at p-value $\le$ 0.05 over paired samples t-test, "**" marks the methods that improve the best baselines at p-value $\le$ 0.01. (ST: Single task for each model, MT: Multi-task trained at the same time for each model). 

\indent In Table \ref {EvaluationTable}, the proposed BMoE achieves significant improvement compared with different baselines. Nevertheless, the BMoE does not show the best performance for the $\mathrm{SO_2}$ prediction task. It achieves the best-balanced performance between these two gas content simultaneous prediction tasks among those baselines by keeping a highly precise prediction result for the $\mathrm{SO_2}$ contents while achieving relatively better performance on the $\mathrm{H_2S}$ prediction task at the same time. More intuitively, it is hard for these single-task baseline methods to learn these multiple quality variables simultaneously. 

According to our results, these single-task baseline methods will suffer performance loss while they are forced to accomplish multi-tasks at the same time. For instance, the content of $\mathrm{H_2S}$ prediction task is strongly affected by the jointly $\mathrm{SO_2}$ prediction task, while the R2 drops from 0.7442 when the model is trained for single-task separately to 0.4476 and shows great variance. The LSTM-based baseline methods perform better, but they still generally show poorer performance facing multi-tasks, which means that the highly related quality variables prediction tasks affect the other prediction tasks in a bad way, as called negative transfer. Furthermore, back to the Autoencoder-based models, SAE may not perform as well as a single AE, while the GSAE works much better in the meantime. LSTM and its variants like VALSTM and SLSTM are considered to be great improvements as they perform best among these single-task learning baselines. These phenomena indicate that the aforementioned reactions will influence the data-driven model performance, and the simultaneous prediction of two gases should consider the reactions to avoid the negative transfer, which makes the model deteriorate. 

\begin{figure}[!t]
\centerline{\includegraphics[width=\columnwidth]{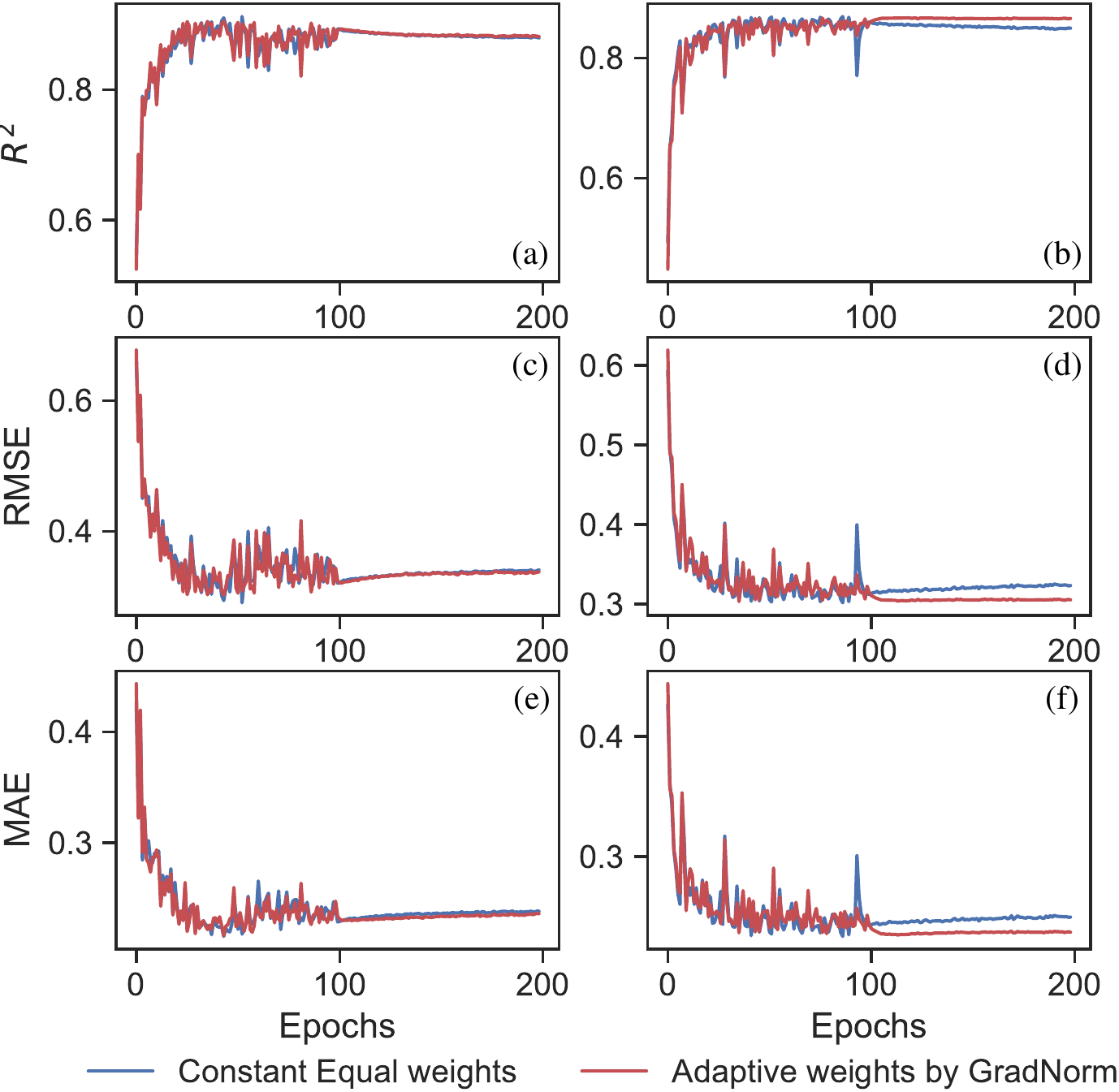}}
\vspace{-0.3cm}
\caption{
The performance of the GradNorm block during the training process. (a)(c)(e) for $\mathrm{H_2S}$ prediction in R2, RMSE, and MAE respectively; (b)(d)(f) for $\mathrm{SO_2}$ prediction in R2, RMSE, and MAE respectively.}
\vspace{-0.6cm}
\label{fig5}
\end{figure}

\subsection{The effectiveness of the GradNorm block.}
\indent In this subsection, the effectiveness of the GradNorm block is discussed in detail. The contribution of the GradNorm block is studied via extracting the training process of the model with constant and equal weights. Meanwhile, the adaptive weights are adjusted by the GradNorm at the same time.
Note that, the learning rate of the training process is dropped as 1/10 of the initial learning rate after 100 epochs.

\indent In Fig. \ref{fig5}, the performance of the quality prediction tasks on $\mathrm{H_2S}$ and $\mathrm{SO_2}$ indicated by the RMSE, R2, and MAE respectively is presented. According to Fig. \ref{fig5}, as the $\mathrm{H_2S}$ prediction task is conducted similarly, with GradNorm block, the $\mathrm{SO_2}$ prediction task performs much better, the R2 score of which raises from 0.8286 to 0.8665, which means that the seesaw phenomenon between multi-tasks has been overcome by the GradNorm block. To better learn how the GradNorm block works in the training process, dynamic change of task weights started by different initial weight ratios is studied as shown in Fig. \ref{fig6}. In such training processes, only the initial task weight ratio is changed from 0.5:1.5 to 1.5:0.5 (weight of $H_2S$ prediction task loss: weight of $SO_2$ prediction task loss). It is obvious that according to the result, no matter what the initial task weight ratio is, the weights are always consistent at nearly 1.5 for $\mathrm{SO_2}$ and 0.5 for $\mathrm{H_2S}$ after the training process. This consistency proves the task balance ability of the GradNorm block.

\subsection{Parameter Sensitivity Study}
Two critical hyperparameters in BMoE are discussed in this section, learning rate $lr$ and relative inverse training rate coefficient $\alpha$, which strongly influence the performance of BMoE.

In Fig. \ref{fig7}, the influence of $\alpha$ is studied by varying it in the range [0,3]. Generally, BMoE performs well when $\alpha$ is at a relatively low value, while it drops obviously with $\alpha$ increasing. Note that $\alpha=0$ means that the GradNorm block always wants to adjust the norms of backpropagated gradients of each task to become equal at W, the subnet weights. Especially, R2 of $\mathrm{H_2S}$ prediction and $\mathrm{SO_2}$ prediction achieve 0.8665 and 0.8490 respectively, while the R2 of $\mathrm{H_2S}$ prediction reduced from 0.8397 to 0.8000 when the $\alpha$ changes from 2.5 to 3.0. Moreover, the performances of the $\mathrm{H_2S}$ and $\mathrm{SO_2}$ prediction tasks roughly keep the same when $\alpha$ increases. It illustrates that $\alpha$ does not cause drastically preference for any single task, but influences the total performance of the BMoE model. As such, $\alpha$ is generally suggested within the range [0.1, 0.5].

Furthermore, the initial learning rate of BMoE varies from 0.001 to 0.02. Remarkably, as the learning rate increases, BMoE performs better before $lr = 0.01$, and then turns to performs worse after that point. Especially, the R2 of $\mathrm{H_2S}$ prediction and $\mathrm{SO_2}$ prediction achieve 0.8653 and 0.8556 respectively at $lr = 0.01$, while they are 0.8114, 0.8215 at $lr = 0.001$ and 0.8409, 0.8387 at $lr = 0.02$. Another observation is that the performance of the BMoE becomes unstable when $lr$ is at a relatively high value. In that way, $lr$ is generally suggested within range [0.008, 0.012].

\subsection{Running Time}

\begin{figure}[!t]
\centerline{\includegraphics[width=\columnwidth]{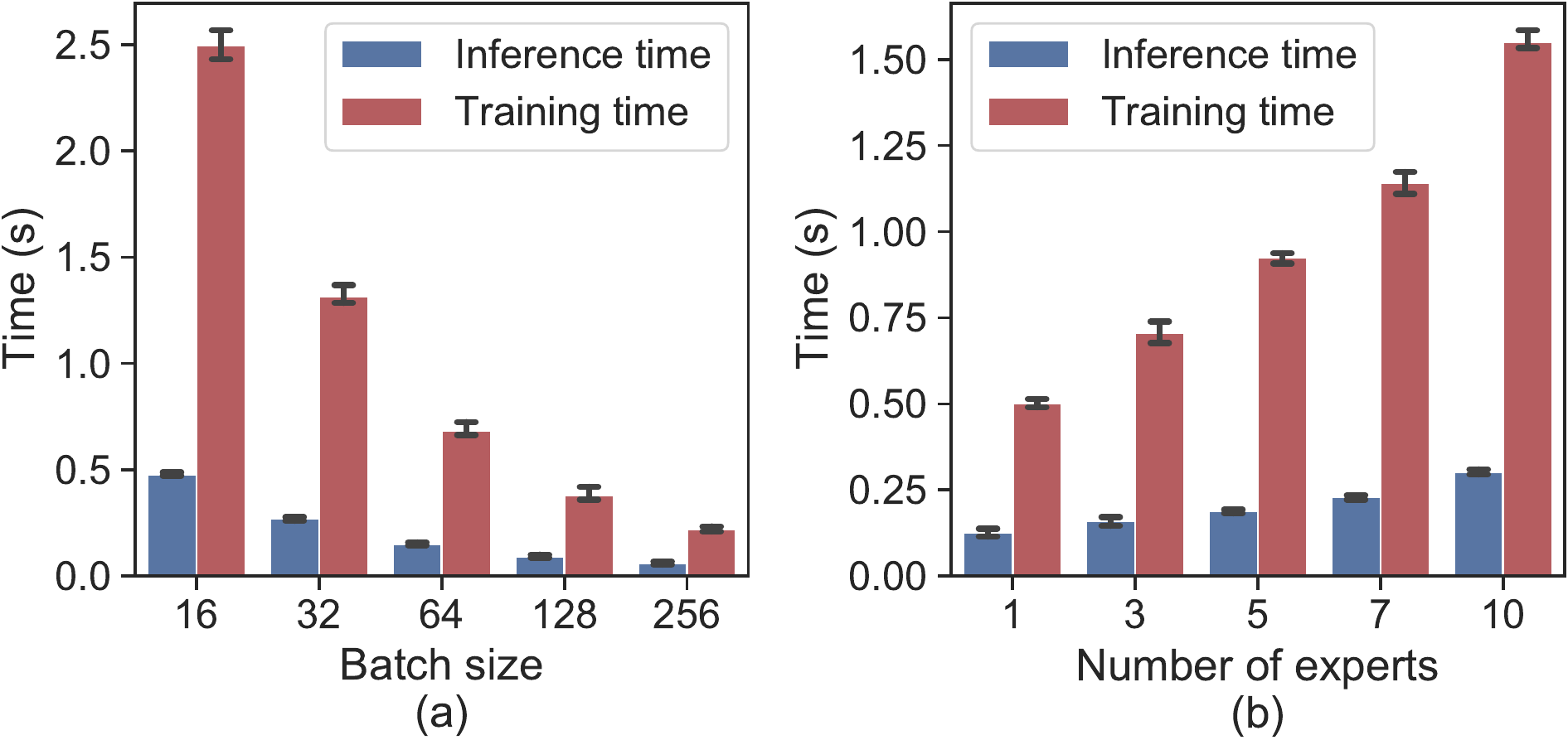}}
\vspace{-0.3cm}
\caption{
The inference time and training time ($\mu\pm3\sigma$) for a single epoch of all the training data (about 6,000 samples) with various (a) batch size and (b) number of experts.}
\vspace{-0.6cm}
\label{fig8}
\end{figure}

The running time of our proposed model is studied in Fig. \ref{fig8}. The results show that the complexity of our proposed model is concentrated on the back propagation process, specifically, the gradnorm algorithm. According to our results, our model can generate prediction result in a relatively high speed, comparing to traditional sensors.

\section{Conclusions}
A novel multi-task learning model combined with MMoE structure and Gradnorm algorithm as a dynamic modeling approach in the soft sensor field called BMoE has been developed in this paper, which can predict multiple variables that are hard to be directly measured in the actual industrial process with good accuracy. As the task imbalance scenario remains one of the main challenges for MTL, the GradNorm algorithm is applied to balance the gradients between different tasks. The experiments on the SRU benchmark show that the proposed BMoE performs better than the widely used autoencoder-based and LSTM-based baseline methods like GSAE, VALSTM, and SLSTM. 

As for future work, more MTL methods like Sub-Network Routing \cite{32}, PLS \cite{33}, and their reasonable application to the soft sensor field is still worth exploring, such models can further improve the efficiency of joint representation learning while helping solve negative transfer problem. Secondly, as samples in the chemical process are generally time series data, a further block that can deal better with time series characteristics needs to be concerned. 

\bibliographystyle{elsarticle-num}
\bibliography{reference}
\end{document}